\documentclass[runningheads]{llncs}
\usepackage{color}
\usepackage{graphicx}
\usepackage[autostyle]{csquotes}
\usepackage{amsmath}
\usepackage{amssymb}
\usepackage{subfig}
\usepackage{tikz}
\usetikzlibrary{automata, positioning}

\begin{document}
\title{A Formulation of Recursive Self-Improvement and Its Possible Efficiency}
\author{Wenyi Wang}
\institute{Computer Science, University of British Columbia \\ \email{wenyw@cs.ubc.ca}}
%


\maketitle              
\begin{abstract}
Recursive self-improving (RSI) systems have been dreamed of since the early days of computer science and artificial intelligence. However, many existing studies on RSI systems remain philosophical, and lacks clear formulation and results.  In this paper, we provide a formal definition for one class of RSI systems, and then demonstrate the existence of computable and efficient RSI systems on a restricted version. We use simulation to empirically show that we achieve logarithmic runtime complexity with respect to the size of the search space, and these results suggest it is possible to achieve an efficient recursive self-improvement.

\keywords{Recursive Self-Improvement  \and Intelligence Explosion \and Stochastic Optimization.}
\end{abstract}

\section{Introduction}
Recursive self-improving systems create new software iteratively. The newly created software should be better at creating future software. With this property, the system has potential to completely rewrite its original implementation, and take completely different approaches \cite{yampolskiy2015seed}. Chalmers' proportionality thesis \cite{chalmers2010singularity} hypothesizes that an increase in the capability of creating future systems proportionally increases the intelligence of the resulting system. With this hypothesis, he shows if a process iteratively generates a greater intelligent system using the current system, then this process leads to a phenomenon many refer to as superintelligence. However, many existing studies of RSI systems remain philosophical or lack clear mathematical formulation or results, e.g. \cite{myhill,nivel2013bounded}. Some mathematically clear work on this topic exist, but they mostly focus on the architectures and methodologies to implement such systems \cite{schmidhuber2003godel,fallenstein2015proof}. Our work is motivated to overcome this weakness by providing a mathematical formulation for a class of RSI procedures. With this formulation, we show that there exist such computable RSI systems. We further study in simulation that this procedure takes logarithmic runtime with respect to the size of search space to find the best program.

\section{The Mathematical Formulation for A Family of RSI Systems}
In this section, we develop a mathematical formulation for a family of RSI systems. To this end, we first examine the necessary elements of an RSI system. An RSI system iteratively improves its current program on the ability to generate \enquote{good} future programs. There are two crucial concepts that should be considered. First, an RSI system can be viewed as a sequence of programs where each program in the sequence generates the next program. Second, each program in the sequence has \textit{increasing} ability to create future programs. Therefore, to define an RSI procedure a set of programs that can generate programs and an order of programs' ability to improve future programs are needed. In the following, we consider a finite search space of programs that generate programs and a total order over it. Notice that a total order over a finite set is isomorphic to a score function. Denote the set of programs by $P$ and the score function by $S$. For convenience, let a lower score represent a higher order. In other words, our objective is to minimize the score function S. Then an RSI system can be described as the following:

\begin{definition}[RSI system]
\textit{Given a finite set of programs $P$ and a score function $S$ over $P$. Initialize $p$ from $P$ to be the system's current program. Repeat until certain criterion satisfied, generate $p' \in P$ using $p$. If $p'$ is better than $p$ according to S, replace $p$ by $p'$.}
\end{definition}

From this definition, one needs to decide how $p \in P$ generates a program.  In general, we should allow the RSI system to generate programs based on the history of the entire process. We assume a simplification that any program in the sequence is independent of all earlier programs given the immediate past program. In other words, the way a program generates a new program is independent of the history, and each program defines a fixed probabilistic distribution over $P$. This procedure defines a homogeneous Markov chain. We will see that even with this restriction, with some score function, the model is able to achieve a desirable runtime performance.

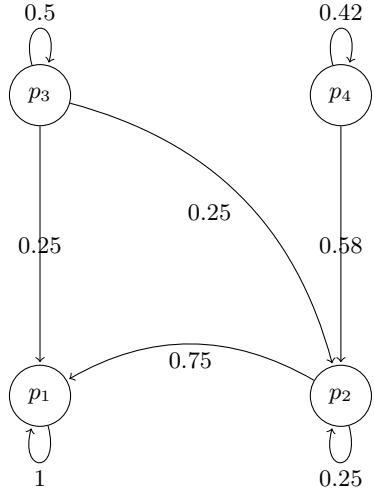
\begin{figure}
	\centering
    \begin{tikzpicture}
        \node[state] at (0,0)            (p1) {$p_1$};
        \node[state] at (4,4) (p4) {$p_4$};
        \node[state] at (0, 4) (p3) {$p_3$};
        \node[state] at (4, 0) (p2) {$p_2$};

        \draw[every loop]
            (p1) edge[loop below] node {1} (p1)
            (p3) edge[loop above] node {0.5} (p3)
            (p3) edge[bend left, auto=right] node {0.25} (p2)
            (p3) edge[] node {0.25} (p1)
            (p2) edge[loop below] node {0.25} (p2)
            (p4) edge[] node {0.58} (p2)
            (p4) edge[loop above] node {0.42} (p4)
            (p2) edge[bend right, auto=left] node{0.75} (p1);
        
    \end{tikzpicture}
    \caption{The Markov chain corresponding to the RSI procedure defined by given scores and program generation probabilities in the example.}
    \label{fig:MC}
\end{figure}

We illustrate the proposed formulation by an example. Consider a set of programs $P=\{p_1, p_2, p_3, p_4\}$ and a score function $S$ over $P$ such that $S(p_i) = i$. According to our formulation, each program can be abstracted as a probabilistic distribution over $P$. To specify the distributions, let $w_i$ be a vector of probabilistic weights of length 4 that represents the probabilistic distribution over $P$ corresponding to $p_i$. In this example we set
\begin{align*}
w_1 &= [0.97, 0.01, 0.01, 0.01],\\
w_2 &= [0.75, 0, 0.25, 0],\\ 
w_3 &= [0.25, 0.25, 0.25, 0.25],\\
w_4 &= [0, 0.58, 0, 0.42].
\end{align*}
Then a possible RSI procedure may do the flowing. It starts from $p_3$. First $p_3$ generates $p_4$. Since $S(p_4)>S(p_3)$, the current program is not updated. Then $p_3$ generates $p_2$. The current program is updated to $p_2$ because $S(p_2)<S(p_3)$. Next $p_2$ generates $p_1$, and the current program updates to $p_1$. Since $p_1$ has the lowest score (highest order), no future program will be updated. Figure \ref{fig:MC} shows the corresponding Markov chain.

\section{The Score Function as Expected Number of Steps}
The last section defines an RSI procedure given a finite set of programs and a score function over it. We have specified the programs, but not the score function. Recall that the score function is to measure the programs' ability to generate \enquote{good} future programs. We assume there is a utility measure being considered that can measure the \enquote{goodness} of a program. This measure need not be the same as the score function. Since the goal of these RSI systems is to find \enquote{good} programs, there needs to be a subset of target programs. Without loss of generality, we can assume there is a unique target program. One can do this because the further analysis will treat the target program as an absorbing state (the state that, once entered, cannot be left) of the Markov process. 

A reasonable utility measure is the expected numbers of steps starting from a program to find the optimal program following our RSI definition. Furthermore, the score function needs to be consistent with the expected numbers of steps from programs to the optimal program following the process defined by itself. We mean that a score function $S$ is consistent if for all $p,p' \in P$, $S(p)>S(p')$ implies that the expected number of steps to reach the optimal program from $p$ is greater than starting from $p'$ . More generally, if one takes some measure for a programs' ability to generate future programs, the score function needs to be consistent with this measure. 

In the following, we describe how to construct a consistent score function. Construct the score function as the expected number of steps to reach the optimal program. To do this, we iteratively update the scores in an nondecreasing order. An intermediate Markov chain always follows the rules of transition defined by the program distributions and current scores. It is obvious that the optimal program should have the minimum score (smaller score represents more preferred program). Initially add the optimal program to the Markov chain, and set its score equals to zero. Set all other programs' score equal infinity. Then repeat until all programs have a finite score. At each step, find program $p$ such that $S(p)=\infty$ and $p$ has the minimum expected number of steps to reach the optimal program. 
Update the score of $p$ as the expected number of steps to reach the optimal program from $p$. The Markov chain is be changed after changing scores. This process of computing the score function can be done in $O(nlogn+m)$ time by dynamic programming, which is similar as the Dijkstra algorithm, where $n$ is the size of programs, and $m$ is the sum of the number of possible programs that each program can generate. We do not describe the efficient way to compute it since the emphasis is the existence and computability of this function.  

Two nice properties hold for this construction. First, the programs are added in an nondecreasing order of scores. Second, the score function equals the expected numbers of steps to reach the optimal program defined by this score function. We will prove the first property. The second property and the consistency of the score function are straightforward from the first property. Before the proof, we describe an example of how such score function is computed given the distributions to generate programs of each program and the optimal program.

Consider the same abstraction of programs as the example in section 2, where $P=\{p_1,p_2,p_3,p_4\}$ with corresponding probabilistic weights
\begin{align*}
w_1 &= [0.97, 0.01, 0.01, 0.01],\\
w_2 &= [0.75, 0, 0.25, 0],\\ 
w_3 &= [0.25, 0.25, 0.25, 0.25],\\
w_4 &= [0, 0.58, 0, 0.42].
\end{align*} Fix $p_1$ to be the optimal program. Initially set $S(p_1)=0$ and $S(p_i)=\infty, i=2,3,4$. The transition function of initial Markov chain is 
\[
\begin{bmatrix}
1 &0 &0 &0 \\
0.75 &0.25 &0 &0\\
0.25 &0 &0.75 &0\\
0 &0 &0 &1
\end{bmatrix}.
\]

At the first step, the expected number of steps from $p_2, p_3, p_4$ following the current Markov chain are $\frac{4}{3}, 4, \infty$. Hence we update $S(p_2)=\frac{4}{3}$. Because of the change of score, transition of the Markov chain change to
\[
\begin{bmatrix}
1 &0 &0 &0 \\
0.75 &0.25 &0 &0\\
0.25 &0.25 &0.5 &0\\
0 &0.58 &0 &0.42
\end{bmatrix}.
\]
Then we compute the expected number of steps from $p_3$ and $p_4$ following the updated Markov chain. By some arithmetic we get the expectation are $\frac{8}{3}$ for $p_3$ and (approximately) $3.057$ for $p_4$. Since $\frac{8}{3}<3.057$, update $S(p_3)=\frac{8}{3}$. By similar procedures, one can compute the score for $S(p_4)$.

\begin{proof}
Let $p_i$ be the $i^{th}$ program being added to the Markov process. Denote the resulting score function by $S$. We need to show that $S(p_i) \leq S(p_{i+1})$ for all feasible $i$'s. 

Prove by induction on $i$. The base case $i=1$ is true since $S(p_1)  = 0$ and $S(p_2) \geq 0$ because it is some expected number of steps. For $i>1$, assume $S(p_j) \leq S(p_{j+1})$ holds for all $j<i$. By definition we know that $S(p_i)$ equals the expected number of steps from $p_i$ to reach $p_1$ following the Markov chain at step $i$. Denote the expected number of steps from $p_{i+1}$ to reach $p_1$ following the Markov chain at step $i$ by E. E satisfies the equation that
\begin{align*}
E = (1-\sum_{k<i} q_{i+1,k})(E+1) + \sum_{k<i} q_{i+1,k} (S(p_k)+1),
\end{align*}
where $q_{i+1,k}$ is the probability that $p_{i+1}$ generates $p_k$. Therefore,
\begin{align*}
E = \frac{1-\sum_{k<i} q_{i+1,k} + \sum_{k<i} q_{i+1,k} (S(p_k)+1)}{\sum_{k<i} q_{i+1,k}}.
\end{align*}
By the constructive process we know that $S(p_i) \leq E$. 

Since $S(p_i) \geq S(p_k)$ for all $k<i$, the Markov chain at step $i+1$ is the Markov chain at step $i$ with some transitions from $p_k$ to $p_i$, where $k<i$. Since for all programs $p_k, k>i$, $S(p_k) = \infty$ at step $i+1$, there is no transition between $p_k$'s for $k>i$. Therefore, similar as $E$,
\begin{align*}
&\quad \, S(p_{i+1}) \\
&= (1-\sum_{k<i} q_{i+1,k} - q_{i+1,i})(S(p_{i+1})+1) + \sum_{k<i} q_{i+1,k} (S(p_k)+1) + q_{i+1,i} (S(p_i)+1),
\end{align*}
and
\begin{align*}
S(p_{i+1}) = \frac{1-\sum_{k<i} q_{i+1,k} + \sum_{k<i} q_{i+1,k} (S(p_k)+1) - q_{i+1,i}S(p_i)}{\sum_{k<i} q_{i+1,k} + q_{i+1,i}}.
\end{align*}
Denote $1-\sum_{k<i} q_{i+1,k} + \sum_{k<i} q_{i+1,k} (S(p_k)+1)$ by $a$ and $\sum_{k<i} q_{i+1,k}$ by $b$. Then $E=\frac{a}{b}$ and $S(p_{i+1}) = \frac{a+q_{i+1,i}S(p_i)}{b+q_{i+1,i}}$. Since $E\geq S(p_i)$, $a \leq S(p_i)b$. Hence $a+S(p_i)q_{i+1,i} \geq S(p_i) (b + q_{i+1,i})$. Thus $S(p_{i+1}) \geq S(p_i)$. $\hfill\square$
\end{proof}

\section{Simulation Results}
\begin{figure}
\centering
\subfloat[Expected numbers of steps]
{
	\includegraphics[width=.5\linewidth]{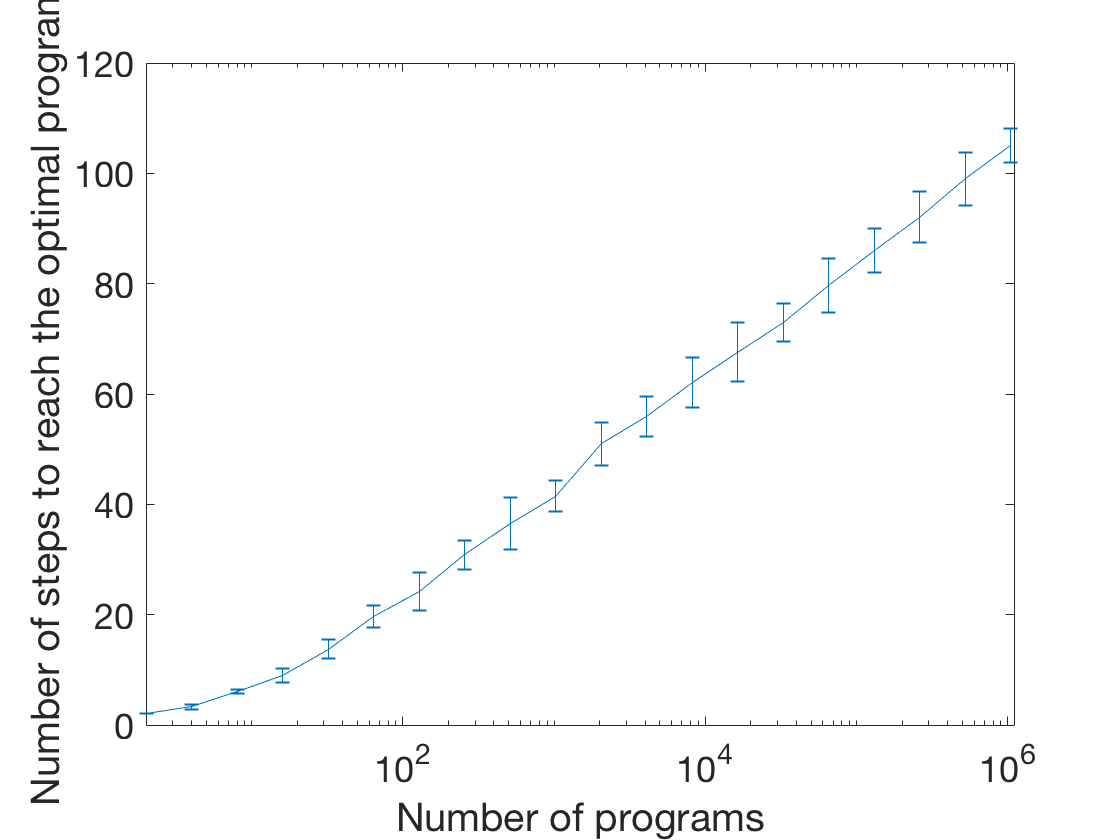}
	\label{fig:es}
}
\subfloat[Ranks]
{
	\includegraphics[width=.5\linewidth]{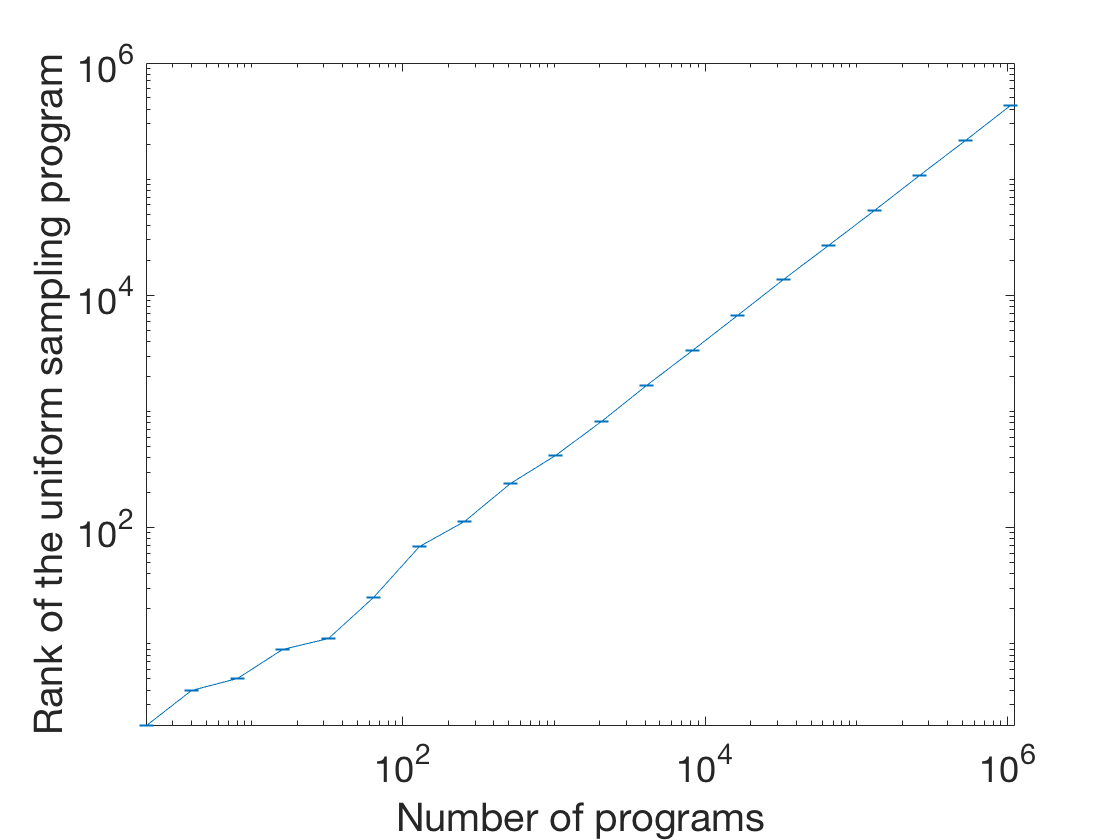}
	\label{fig:rank}
}
\caption{These two figures show simulation results of the expected number of steps to the optimal program (a) and ranks (b) of the program that generates programs uniformly.}
\label{fig:exps}\end{figure}

We test the performance of the proposed RSI procedure in simulation with randomly generated abstraction of programs. For each of the experiments, a fixed number of programs is chosen from $n = 2^l, l=1,2,\dots,20$. The first program is designed to generate programs uniformly over all programs. Other programs generate programs follow a weighted distribution over a subset of programs. The sizes of subsets are drawn i.i.d. from the uniform distribution over integers between 10 and 100. Given the size of a subset, the subset and corresponding weights are drawn uniformly over the feasible supports. With 10 repeats for each $l=1,2,\dots,20$, the expected number of steps for the first program to reach the optimal program and its rank over all programs are shown in figure \ref{fig:exps}. Figure \ref{fig:es} suggests a linear relation between $l$ and  expected number of steps, and figure \ref{fig:rank} suggests a linear relation between $n$ and rank of the first program. A linear regression model fits $l$ and expected number of steps returns an R-squared value equals 0.983, which indicates the linear model can explain a lot of the data. Similarly, the linear regression fit to $n$ and rank of the first program has R-squared value equals 1.0.

\begin{figure}
\centering
\includegraphics[width=.9\linewidth]{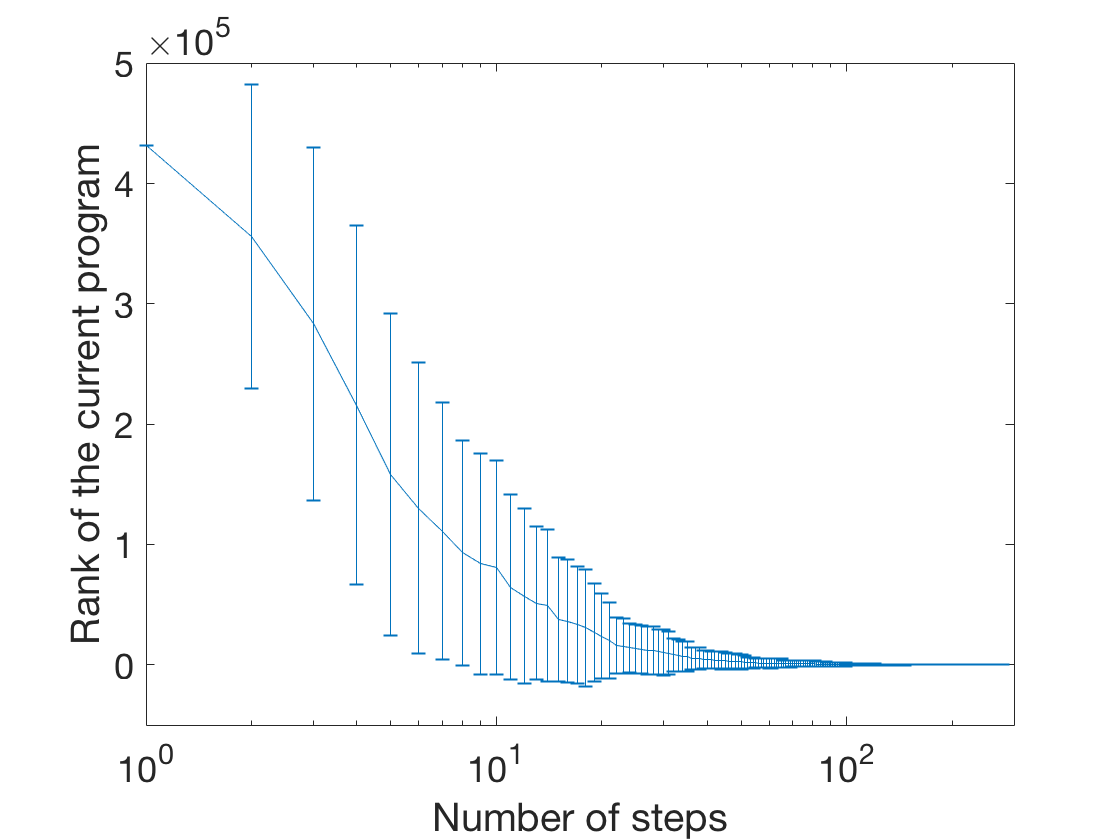}
\caption{A simulation result of running our proposed RSI procedure given the precomputed score function.}
\label{fig:sim}
\end{figure}

For a fixed RSI system with $n=2^{20}$, we run 100 simulations of proposed procedure starting from the first program. Figure \ref{fig:sim} shows an error-bar of the ranks of current program at different number of steps of the simulation. We see that before some of the processes reach the optimal program, the ranks improve exponentially in a statistical sense. All of the processes converge to the global optimal program.

\section{Discussion and Future Works}
In summary, we formulate a family of RSI procedures. For a more restricted family of RSI procedures satisfying the Markov assumption, we prove that a consistent score function exists, and we describe an algorithm to compute it. We study runtime of the restricted systems empirically. Experimental results suggest a logarithmic relation between the runtime and the number of programs. These results suggest a possibility of efficient recursive self-improvement. For future works, one may expand the model by embedding histories when generating a new program. Another possible extension is to model the programs taking a program as argument and return a suggested improvement of the given program.  It is remarkable that in the simulations, the score function is precomputed, which takes more time than enumerate every program to find the optimal. From the practical point of view, to make the proposed procedure applicable, one needs to design an oracle score function, where at each evaluation it dose not need to process all other programs. One possible approach is to let each program take an program design task that can be evaluated as argument, and evaluate a program based on its performance on the evaluable tasks. The more rigorous approach is to study the reasoning of a program of its future behaviour including rewrites. This phenomenon is referred as Vingean reflection in Fallenstein and Soares's work \cite{fallenstein2015vingean}. Alternatively, Steunebrink and Schmidhuber formulate it as a proof finding problem \cite{steunebrink2012towards}. On the high level, this problem remains open and challenging. Since a practical score function may not have the desired properties as we analyzed in the ideal case, it would be interesting to study the behaviour of proposed procedures when the score function is biased, noisy or inconsistent.

 \bibliographystyle{splncs04}
 \bibliography{mybibliography}
\end{document}